\documentclass[10pt,twocolumn,letterpaper]{article}

\usepackage[pagenumbers]{iccv} 

\definecolor{iccvblue}{rgb}{0.21,0.49,0.74}
\usepackage[pagebackref,breaklinks,colorlinks,allcolors=iccvblue]{hyperref}
\usepackage{subcaption}
\usepackage{multirow}
\usepackage[ruled]{algorithm2e}
\usepackage{caption}

\usepackage{booktabs}
\PassOptionsToPackage{table}{xcolor}
\usepackage[pagenumbers]{iccv}
\usepackage{colortbl}
\usepackage{pifont}
\newcommand{\cmark}{\ding{51}}

\usepackage{graphicx} 
\usepackage{tabularx}
\newcolumntype{C}{>{\centering\arraybackslash}X}

\definecolor{lightgray}{gray}{0.9}
\definecolor{lightpink}{rgb}{1.0,0.9,0.9}

\def\paperID{****} 
\def\confName{ICCV}
\def\confYear{2025}

\title{LoRA-Loop: Closing the Synthetic Replay Cycle for Continual VLM Learning}

\author{Kaihong Wang\thanks{Work done at Boston University}\\
Waymo\\
{\tt\small kaihongwang@google.com}
\and
Donghyun Kim\\
Korea University\\
{\tt\small d\_kim@korea.ac.kr}
\and
Margrit Betke\\
Boston University\\
{\tt\small betke@bu.edu}
}

\begin{document}

\maketitle

\begin{abstract}
Continual learning for vision–language models has achieved remarkable performance through synthetic replay, where samples are generated using Stable Diffusion to regularize during finetuning and retain knowledge. However, real‐world downstream applications often exhibit domain‐specific nuances and fine‐grained semantics not captured by generators, causing synthetic‐replay methods to produce misaligned samples that misguide finetuning and undermine retention of prior knowledge. In this work, we propose a LoRA-enhanced synthetic-replay framework that injects task-specific low-rank adapters into a frozen Stable Diffusion model, efficiently capturing each new task’s unique visual and semantic patterns. Specifically, we introduce a two-stage, confidence-based sample selection: we first rank real task data by post-finetuning VLM confidence to focus LoRA finetuning on the most representative examples, then generate synthetic samples and again select them by confidence for distillation. Our approach integrates seamlessly with existing replay pipelines—simply swap in the adapted generator to boost replay fidelity. Extensive experiments on the Multi-domain Task Incremental Learning (MTIL) benchmark show that our method outperforms previous synthetic-replay techniques, achieving an optimal balance among plasticity, stability, and zero-shot capability. These results demonstrate the effectiveness of generator adaptation via LoRA for robust continual learning in VLMs.
\end{abstract}

\section{Introduction}
\label{sec:intro}
Vision–language models (VLMs) have seen remarkable advances in recent years. Representative architectures such as CLIP~\cite{RadfordKiHaRaGoAgSaAsMi21} learn a joint embedding space between images and text via a contrastive objective on massive, diverse datasets (e.g., ALIGN~\cite{JiaYXCPPLSLD21} and Florence~\cite{YuanChChCoDaGa21}), enabling strong generalization across downstream tasks. In particular, their use of contrastive loss and inclusion of billions of image–text pairs enable strong zero‐shot transfer capabilities. Nevertheless, no pretraining data can cover every visual domain, so VLMs often require further finetuning on specific labeled datasets to acquire task‐relevant knowledge without forgetting previous knowledge.
Continual learning (CL) addresses this by updating models on new tasks while preserving earlier knowledge. In classical CL, \textit{plasticity} means integrating new, task-relevant information, and \textit{stability} means retaining performance on all previously seen task data (i.e., avoiding catastrophic forgetting). For VLMs, stability must also encompass preserving their zero-shot generalizability to novel classes from different tasks and domains learned during pretraining, so that tuning on one task does not degrade the model's ability to recognize unseen concepts.
\begin{figure}[t]
  \centering
   \includegraphics[width=\linewidth]{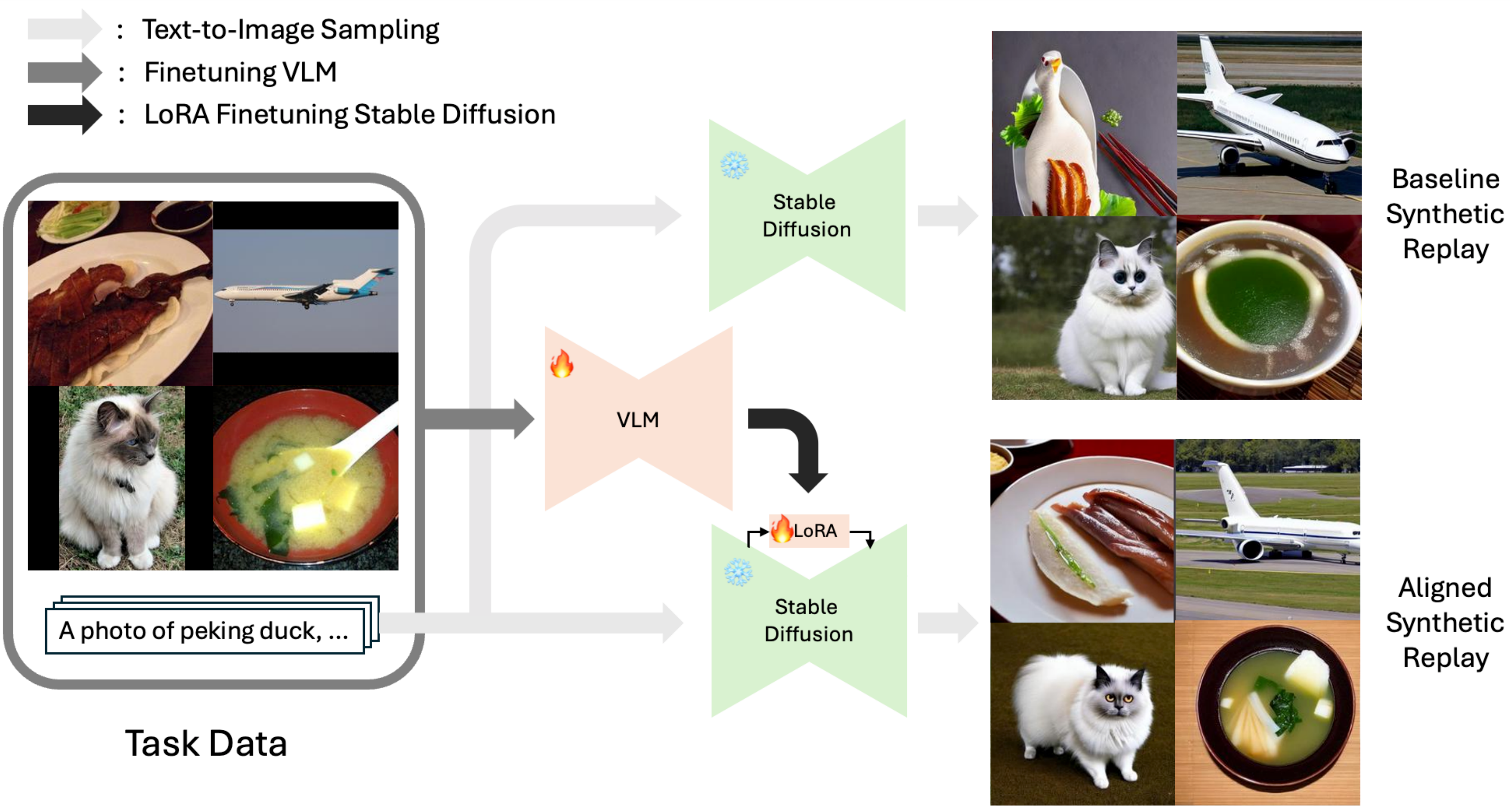}
    \vspace{-0.8cm}
   \caption{Comparison of baseline synthetic replay (top) versus our LoRA‐Loop (bottom). By learning from the feedback from VLM finetuning and aligning generated samples with task data, we boost the fidelity of synthetic replay data and improve distillation quality to better retain existing knowledge. }
    \vspace{-0.6cm}
   \label{fig:teaser}
\end{figure}
Zheng et al.'s ZSCL~\cite{zheng2023preventing} first tackled CL for VLMs under multi-domain settings by distilling on large, diverse reference data (e.g., ImageNet~\cite{russakovsky2015imagenet} or CIFAR~\cite{krizhevsky2009learning}), aligning post-finetuning activations with their original values to prevent drift. They further proposed the more challenging and realistic Multi-domain Task Incremental Learning (MTIL) benchmark to evaluate performance across varied domains.
More recently, GIFT~\cite{WuShWa25} leverages a frozen Stable Diffusion~(SD) model~\cite{RoBlaLoEsOm22} to generate synthetic “reference” images on the fly, instead of storing a large real dataset. This approach cuts storage costs and enables concept rehearsal without any generator finetuning.

Despite their success in preserving VLMs' zero-shot generalizability on broad domains, we observe that synthetic-replay methods falter when the downstream tasks lie outside the pretraining distribution.  Although Stable Diffusion's outputs generally align well with the pretraining data of VLMs, they struggle to capture the fine-grained semantics and domain-specific characteristics of specialized tasks.  For instance, in an aircraft‐model recognition task—distinguishing a Boeing 737-800 from a 727-200—Stable Diffusion might frequently lack the nuanced semantic grounding to render the subtle tail shapes, engine placements, or fuselage contours that differentiate these variants. Besides, domain and distributional shifts often arise when the task dataset is collected in a particular subdomain, such as military versus commercial aviation imagery. These combined semantic and domain shifts undermine the fidelity of replay, limiting both the retention of prior-task knowledge and the continued zero-shot transfer ability of the VLM.
A common remedy is to retain a small buffer of real task examples for replay to bridge domain and semantic gaps. However, this conflicts with privacy constraints and presents a difficult trade-off: too few samples impair replay effectiveness, while larger buffers inflate storage costs and invite overfitting, hurting the generalizability of VLMs.

To bridge these gaps efficiently, we introduce task-specific LoRA adapters~\cite{HuSWALWWC22} into the Stable Diffusion pipeline. Specifically, since a large number of classes learned in finetuning must be replayed, we first select a few representative real examples per class by ranking post-training confidence in the VLM. This focuses LoRA's limited adaptation capacity on the most informative samples. We then inject low-rank weight updates into Stable Diffusion and finetune only those adapters on the selected task data, capturing the task's unique visual and semantic patterns while leaving the base model, and its broad zero-shot priors, untouched. For future synthetic replay, we generate a pool of candidate images with the LoRA-adapted diffusion model and again rank them by their VLM confidence, selecting only the well-aligned samples for distillation. This two-stage, confidence-based selection balances \textit{stability} with \textit{plasticity}, ensuring that replay sets faithfully reflect both the distributional nuances and semantic detail of each task. Our LoRA-enhanced replay seamlessly integrates with existing baselines like GIFT, simply swapping in the adapted generator to boost replay quality without altering the overall framework.
To summarize, the contributions of this work include:
\begin{itemize}
\item \textbf{LoRA-driven feedback loop}: We close the loop from VLM finetuning back to diffusion synthesis by training task-specific LoRA adapters that bridge domain and semantic gaps in synthetic replay.
\item \textbf{Confidence-based exemplar selection}: We introduce a two-stage criterion, first on real data, then on synthetic samples, to 
ensure the alignment of samples to task data for effective distillation in CL for VLMs.
\item \textbf{Competitive performance}: On MTIL benchmarks, our LoRA-enhanced synthetic-replay outperforms prior methods, striking a superior balance among plasticity, stability, and zero-shot generalizability.
\end{itemize}

\section{Related Works}
\label{sec:relatedworks}

\subsection{Vision-Language Models}
\label{sec:relatedworks_vlm}

Vision–language models (VLMs)~\cite{RadfordKiHaRaGoAgSaAsMi21,JiaYXCPPLSLD21,YuanChChCoDaGa21,ZhaiWMSK0B22,YuWVYSW22} use large‐scale contrastive pretraining on massive image–text pairs to align visual and textual representations, achieving state‐of‐the‐art zero‐shot transfer and strong generalization. However, task‐specific applications still require finetuning without eroding the pretrained knowledge. Parameter‐efficient methods include CLIP‐Adapter~\cite{GaoGZMFZLQ24}, which trains a lightweight adapter head on frozen CLIP features. Prompt‐based approaches (CoOp~\cite{ZhouYLL22}, CoCoOp~\cite{ZhouYL022}, VPT~\cite{JiaTCCBHL22}) learn a small set of continuous prompt tokens while keeping the backbone frozen. LoRA~\cite{HuSWALWWC22} injects low‐rank adaptation matrices into each transformer layer and trains only the added weights.

\subsection{Continual Learning}
\label{sec:relatedworks_cl}

Continual learning (CL) aims to sequentially learn new tasks while preserving performance on previous ones without accessing their original data. Memory-replay methods~\cite{LavdaRa18,Lopez-PazR17,PrabhuTD20,RebuffiKSL17,ShinLKK17} keep a small buffer of past examples for rehearsal, trading off storage and privacy for effectiveness. Regularization-based approaches~\cite{KirkpatrickPRVD16,LiH16,LeeKJHZ17,ZenkePG17,AljundiBERT18,DharSPWC19,HouPLWL19,DouillardCORV20} regularize the model by adding a penalty on changes to parameters deemed important for earlier tasks, thus retaining prior knowledge but potentially limiting flexibility on new tasks. Dynamic-architecture techniques~\cite{AljundiCT17,YoonYLH18,YanX021,DouillardRCC22,HuLLGV23,YeBo23} allocate task-specific modules or expand model capacity, which mitigates forgetting at the cost of increased complexity and limited parameter sharing.
For vision–language models, continual learning carries the extra requirement of maintaining zero-shot generalizability. 
Prior work divides into two main categories: robust backbone adaptation~\cite{WortsmanILKHFNS21,GoyalKGKR23,HuangLSZWLDTY24} as well as parameter-efficient task modules~\cite{WangHH22,WangZESZLRSPDP22,WangZLZSRSPDP22,ZhouZWNYZL25}.
Zheng et. al~\cite{ZhengMWQYY23} first introduce a multi-domain continual learning benchmark and use distillation on a large reference dataset to preserve zero-shot capacity, while MoE-Adapter~\cite{YuZ0H0LH24} integrates Mixture-of-Experts adapters to capture new-task knowledge without degrading the model's general capabilities. 

\subsection{Learning from Synthetic Data}
\label{sec:relatedworks_syn}
With advances in generative models, synthetic data has become a valuable resource for training discriminative models. Early work explored representation learning from generated samples~\cite{TianFaIsChKr23,FanCKKIT24,TianFaChKaKrIs24,SinghNaHoScRo24} and leveraged synthetic images or captions to boost VLM performance, especially in retrival tasks~\cite{LiuHLY23,NguyenGIOS23,LiTHWZXRMLZZX24,ChenLDZHWZL24,HammoudItPiToBiGh24,LiuLHXYSCHCZ24}. 
More recently, synthetic replay has been applied to continual learning, generating future rehearsal examples~\cite{GaoL23a,JodeletLPM23,KimCKTB24,MengZYZZW24}. GIFT~\cite{WuShWa25} takes this further by introducing Stable Diffusion to synthesize images for VLM continual learning but assumes perfect alignment between generated and real task data, overlooking inherent domain/semantic gaps. In contrast, we introduce a feedback-driven mechanism that guides the generator to produce higher-fidelity samples tailored for effective replay.

Outside CL, several closed-loop methods have leveraged feedback from discriminative models to steer diffusion-based data synthesis.
Askari-Hemmat et al.~\cite{HemmatPBDR24} incorporate classifier‐derived signals into latent diffusion sampling to oversample hard or underrepresented classes to mitigate long-tail imbalances. 
Yeo et al.~\cite{YeoABARAZ25} optimize continuous prompt embeddings with classifier gradients to craft adversarial prompts, guiding the diffusion process toward more challenging, task-aligned examples.  
In contrast, we employ task-specific LoRA adapters to finetune the diffusion generator itself, offering a more straightforward and efficient mechanism to align synthetic replay samples with task domains for VLM continual learning.

\begin{algorithm}[t!]
\caption{LoRA-Loop: Synthetic Replay for Continual VLM Learning}
\label{alg:lora_loop_topk}
\SetKwInOut{Input}{Input}
\SetKwInOut{Output}{Output}
\SetKwComment{Comment}{// }{}
\Input{
  Pretrained VLM $f^0$, base generator $G_{\phi}$,\\
  task data $\{(X^i,Y^i,C^i)\}_{i=1}^n$, \\
  base-class pool $C_0$, Prompt template $T$, \\
  sample budget $M_{\rm pre}$, sample selection $k$,\\
  LoRA selection $l$\\
}
\Output{Finetuned VLM $f^n$}
\BlankLine
$\mathcal A \leftarrow \{\}$ \tcp{No adapters initially}

$C \leftarrow C_0$ \tcp{Init. by ImageNet class}

\For{$i\leftarrow1$ \KwTo $n$}{
  \CommentSty{/* 1.Sample synthetic replay*/}
  
  $S \leftarrow \emptyset$\;
  \ForEach{$c\in C$}{
    \eIf{$\exists\,(A_j,C^j)\in\mathcal A:\;c\in C^j$}{
      \tcp{choose LoRA adapter}
      $\tilde G\leftarrow G_{\phi + A_j}$\;
    }{
      \tcp{Use SD for base class}
      $\tilde G\leftarrow G_{\phi}$\;
    }
    $S_{\rm cand}\leftarrow\emptyset$\;
    \For{$m\leftarrow1$ \KwTo $M_{\rm pre}$}{
      $p\leftarrow T(c)$\;
      $x\leftarrow \tilde G.\mathrm{generate}(p,\mathrm{seed})$\;
      $S_{\rm cand}\leftarrow S_{\rm cand}\cup\{(x,p)\}$\;
    }
    \tcp{Filter by CLIP's score}
    $S\leftarrow S\;\cup\;\mathrm{SampleTopK}\bigl(S_{\rm cand},k,\;f^{\,i-1}\bigr)$\;
  }
  \BlankLine

  \CommentSty{/* 2.Finetuning via GIFT*/}
  
  $\mathcal L \leftarrow \mathrm{ComputeGIFTLoss}(f^{\,i-1}, X^i, Y^i, S)$\;
  
  $f^i\leftarrow\mathrm{Optimize}(f^{\,i-1},\mathcal L)$\;
  \BlankLine
  \CommentSty{/* 3.LoRA finetuning SD*/}
  
  $D_{\rm lora}\leftarrow\mathrm{SelectLoRAData}(f^i,X^i,Y^i,l)$\;
  $A_i\leftarrow\mathrm{LoRA\_Finetune}(G_{\phi},D_{\rm lora})$\;

  $\mathcal A \leftarrow \mathcal A \cup \{(A_i,\,C^i)\}$ ;\
  \BlankLine
  
  \CommentSty{/* 4.Expand class pool*/}
  
  $C\leftarrow C\cup C^i$\;
}
\end{algorithm}

\section{Methodology}
\label{sec:method}

\subsection{Preliminaries}
\label{sec:prelim}

\noindent\textbf{Continual Learning.}
Given $n$ tasks $\{\mathcal{T}^1,\dots,\mathcal{T}^n\}$, continual training proceeds sequentially on each task
$\mathcal{T}^i=(\mathcal{D}^i,\mathcal{C}^i)$, where
the dataset $\mathcal{D}^i=\{(x_j^i,y_j^i)\}_{j=1}^{N_i}$ with images $x_j^i$ and one-hot labels $y_j^i\in\{0,1\}^{m_i}$, and the class set
$\mathcal{C}^i=\{c_j^i\}_{j=1}^{m_i}$, with $m_i = |\mathcal{C}^i|$ the number of classes in task \(\mathcal{T}^i\).  In \emph{task-incremental learning}, the task identity $t$ is known at inference, so the model classifies over $\mathcal{C}^t$, whereas in \emph{class-incremental learning} it predicts over the unified set $\bigcup_{i=1}^n\mathcal{C}^i$.

\noindent\textbf{Vision–Language Model.}
This paper focuses on Contrastive Language–Image Pretraining (CLIP)~\cite{RadfordKiHaRaGoAgSaAsMi21} as the backbone VLM. During pretraining, CLIP jointly learns an image encoder $f_i(\cdot)$ and a text encoder $f_t(\cdot)$. Given an input image $x$, the probability of class $y_i$ is computed as:
\begin{equation}\label{eq:clip-prob}
  p(y_i \mid x) = \frac{\exp\bigl(\cos(z, w_i)/\tau\bigr)}{\sum_{j=1}^{|\mathcal{Y}|} \exp\bigl(\cos(z, w_j)/\tau\bigr)},
\end{equation}
where $z = f_i(x)$ is the image embedding, $w_i = f_t(t_i)$ is the text embedding of the prompt $t_i$ (e.g., ``a photo of a $\{c_i\}$''), $\cos(\cdot,\cdot)$ denotes cosine similarity, and $\tau$ is a learnable temperature. For downstream tasks, we finetune CLIP using the cross-entropy loss over the ground-truth labels.

\noindent\textbf{Low-Rank Adaptation (LoRA).}
LoRA~\cite{HuSWALWWC22} is a parameter-efficient finetuning method that injects low-rank adapters into each frozen weight matrix of a pretrained model. Given a base weight $W_0\in\mathbb{R}^{d\times d}$, LoRA represents the task-specific update as $\Delta W = A\,B$, with $A\in\mathbb{R}^{d\times r}$ and $B\in\mathbb{R}^{r\times d}$, where $r\ll d$. The adapted weight becomes $W' = W_0 + \Delta W$. 

\subsection{Overview of LoRA-Loop}
Building on the GIFT framework \cite{WuShWa25}, our LoRA–Loop closes the feedback loop from VLM finetuning to the diffusion generator, enabling per-task domain adaptation. The detailed process is illustrated in Algorithm~\ref{alg:lora_loop_topk}. We begin by initializing the replay class pool to the ImageNet classes. At task \(i\), we sample a compact, high-quality replay set to ensure the generated samples better align with previous task data for effective replay and distillation (Step 1 in Algorithm~\ref{alg:lora_loop_topk}, Sec.~\ref{sec:sample_sd}). We then update the VLM using the original GIFT losses on both the current task data and the synthetic replay. Finally, we select a balanced mix of prototypical and boundary examples from the task data as representative training inputs for LoRA adapter finetuning, producing a domain-specialized generator for future replay (Step 3 in Algorithm~\ref{alg:lora_loop_topk}, Sec.~\ref{sec:lora_sd}).

\subsubsection{Synthetic Replay Sample Filtering for Distillation}
\label{sec:sample_sd}
To obtain the replay set $S$ at task $i$, for each class $c\in C$ we choose either the base generator $G_\phi$ or its adapted variant $G_{\phi+A_j}$ (if $c\in C^j$ for some $(A_j,C^j)\in\mathcal A$) as the generator $\tilde G$, and generate $M_{\rm pre}$ candidates via 
$$
(x_j,p_j)\sim \tilde G.\mathrm{generate}\bigl(T(c)\bigr), \qquad j = 1, \dots, M_{\rm pre}
$$
We then score each pair by computing the confidence $\mathrm{conf}_j$ via the frozen VLM from the last round $f^{i-1}$:
$$
\mathrm{conf}_j = \cos\bigl(f_{\rm img}^{\,i-1}(x_j),\,f_{\rm txt}^{\,i-1}(p_j)\bigr).
$$ 
and sort $\mathrm{conf}_j$ to retain the top‐$k$ pairs for each class to form $S$ instead of selecting a specific confidence threshold. This ensures $S$ contains the samples best aligned with the domain of the previous task data, improving distillation efficiency while controlling memory and compute.

\subsubsection{LoRA Finetuning for Stable Diffusion}
\label{sec:lora_sd}
After obtaining the updated VLM $f^i$ via GIFT losses, we measure the confidence of each training example in the current task data $(x_j,y_j)\in (X^i,Y^i)$ by computing confidence on the frozen VLM $f^i$ :
$$
\mathrm{conf}_j = \cos\bigl(f_{\rm img}^i(x_j),\,f_{\rm txt}^i(T(y_j))\bigr).
$$ 
For each class in $C^i$, we select the $l$ examples, half with the highest $\mathrm{conf}_j$, representing the most prototypical samples, and the other half with the lowest $\mathrm{conf}_j$, representing the edge cases, to form the balanced set $D_{\rm sel}$. A LoRA adapter $A_i$ with a rank of $r$ is finetuned on $D_{\rm sel}$ and stored, yielding the domain‐specialized generator $G_{\phi+A_i}$ for future sampling of classes in $C^i$.

\section{Experiments}
\label{sec:experiment}

\subsection{Experiment Settings}
\label{sec:exp_settings}

\noindent\textbf{Datasets.}
We evaluate our approach multi‐domain task‐incremental learning (MTIL). MTIL is particularly challenging, including 11 datasets and a total of 1,201 classes across various domains: Aircraft~\cite{MajiRKBV13}, Caltech101~\cite{LiFP07}, CIFAR100~\cite{krizhevsky2009learning}, DTD~\cite{CimpoiMKMV14}, EuroSAT~\cite{HelberBDB19}, Flowers~\cite{NilsbackZ08}, Food~\cite{BossardGG14}, MNIST~\cite{Deng12}, OxfordPet~\cite{ParkhiVZJ12}, StanfordCars~\cite{Krause0DF13}, and SUN397~\cite{XiaoHEOT10}. We follow the two‐order training protocol from ZSCL~\cite{zheng2023preventing}, performing ablations on the default MTIL order I.

\noindent\textbf{Evaluation Metrics.}
We adopt the “Transfer”, “Last”, and “Avg.” metrics introduced in ZSCL~\cite{zheng2023preventing}. “Transfer” quantifies the model’s zero‐shot performance on unseen task data and its retention of pretraining knowledge, while “Last” measures how well the model preserves downstream task performance over time. “Avg.” computes the mean of all performance during the entire finetuning process on a task, capturing the stability–plasticity trade‐off.

\noindent\textbf{Implementation Details.}
We build on prior continual VLM learning work~\cite{zheng2023preventing,WuShWa25} using CLIP with a ViT‐B/16 backbone~\cite{DosovitskiyB0WZ21}.  Each task is finetuned for $1{,}000$ iterations with a batch size of $64$.  For synthetic replay, at each task we draw $M_{\rm pre}=8$ candidates per class using Stable Diffusion v1.5~\cite{RoBlaLoEsOm22} (classifier‐free guidance scale $7.5$, $50$ denoising steps) and retain the top-$k=1$ images per class based on CLIP cosine similarity.  After updating the VLM to $f^i$, we score all training examples and select $l=2$ samples per class to form the LoRA training set.  We then finetune a rank-$r=4$ LoRA adapter on Stable Diffusion for $100$ epochs using AdamW~\cite{LoshchilovH19} with a learning rate of $1\times10^{-4}$, $(\beta_1,\beta_2)=(0.9,0.999)$, and weight decay $1\times10^{-2}$.


\subsection{Results}
\label{sec:exp_results}
\begin{table}[t!]
  \centering
  \caption{Comparison of SOTA methods on MTIL Order I. $^*$ indicates reproduced results.}
  \label{tab:mtil-order1}
  \resizebox{\columnwidth}{!}{%
  \begin{tabular}{l  c  c  c  c  c  c}
    \toprule
    Method & Transfer & $\Delta$ & Avg.\ & $\Delta$ & Last & $\Delta$ \\
    \midrule
    Zero-shot              & 69.4 & —      & 65.3 & —      & 65.3 & —      \\
    Continual Finetune     & 44.6 & —      & 55.9 & —      & 77.3 & —      \\
    \midrule
    $\ell_2$ baseline     & 61.0 & 0.0    & 62.7 & 0.0    & 75.9 & 0.0    \\
    \midrule
    WiSE-FT~\cite{WortsmanILKHFNS21}      & 52.3 & \textcolor{blue}{-8.7}  & 60.7 & \textcolor{blue}{-2.0} & 77.7 & \textcolor{red}{+1.8} \\
    ZSCL~\cite{zheng2023preventing}         & 68.1 & \textcolor{red}{+7.1}   & 75.4 & \textcolor{red}{+12.7} & 83.6 & \textcolor{red}{+7.7} \\
    MoE-Adapter~\cite{YuZ0H0LH24}  & 68.9 & \textcolor{red}{+7.9}   & 76.7 & \textcolor{red}{+14.0} & 85.0 & \textcolor{red}{+9.1} \\
    GIFT$^*$~\cite{WuShWa25}            & 69.7 & \textcolor{red}{+8.7} & 77.3 & \textcolor{red}{+14.6} & 85.4 & \textcolor{red}{+9.5} \\
    LoRA-Loop (Ours)            & 69.8 & \textbf{+8.8} & 77.6 & \textbf{+14.9} & 86.0 & \textbf{+10.1} \\
    \bottomrule
  \end{tabular}
  }
\end{table}
\begin{table}[t!]
  \centering
  \caption{Comparison of SOTA methods on MTIL Order II.  $^*$ indicates reproduced results.}
  \label{tab:mtil-order2}
  \resizebox{\columnwidth}{!}{%
  \begin{tabular}{l  c  c  c  c  c  c}
    \toprule
    Method & Transfer & $\Delta$ & Avg.\ & $\Delta$ & Last & $\Delta$ \\
    \midrule
    Zero-shot              & 65.4 & —      & 65.3 & —      & 65.3 & —      \\
    Continual Finetune     & 46.6 & —      & 56.2 & —      & 67.4 & —      \\
    \midrule
    $\ell_2$ baseline     & 60.6 & 0.0    & 68.8 & 0.0    & 77.2 & 0.0    \\
    \midrule
    WiSE-FT~\cite{WortsmanILKHFNS21}      & 51.0 & \textcolor{blue}{-9.6}  & 61.5 & \textcolor{blue}{-7.3} & 72.2 & \textcolor{blue}{-5.0} \\
    ZSCL~\cite{zheng2023preventing}         & 64.2 & \textcolor{red}{+3.6}   & 74.5 & \textcolor{red}{+5.7}  & 83.4 & \textcolor{red}{+6.2} \\
    MoE-Adapter~\cite{YuZ0H0LH24}  & 64.3 & \textcolor{red}{+3.7}   & 74.7 & \textcolor{red}{+5.9}  & 84.1 & \textcolor{red}{+6.9} \\
    GIFT$^*$~\cite{WuShWa25}            & 66.1 & \textcolor{red}{+5.5} & 75.8 & \textcolor{red}{+7.0}  & 85.2 & \textcolor{red}{+8.0} \\
    LoRA-Loop (Ours)            & 66.3& \textbf{+5.7} & 75.9 & \textbf{+7.1}  & 85.5 & \textbf{+8.3} \\
    \bottomrule
  \end{tabular}
  }
\end{table}

\subsubsection{Comparison To Baselines}
\label{sec:comparison}
We evaluate our method on the two MTIL benchmarks (Order I and II) and report results in Tab.~\ref{tab:mtil-order1}, Tab.~\ref{tab:mtil-order2}, and more detailed scores in our appendix.  As references, we include: (1) the zero‐shot CLIP backbone (no finetuning); (2) Continual Finetuning, which sequentially finetunes CLIP on each task without any continual learning mechanism; (3) an $\ell_2$ regularization baseline that constrains parameter drift back toward the pretrained weights; and (4) recent continual‐VLM methods WiSE‐FT~\cite{WortsmanILKHFNS21}, ZSCL~\cite{zheng2023preventing}, MoE‐Adapter~\cite{YuZ0H0LH24}, and GIFT~\cite{WuShWa25}.

As shown in Table~\ref{tab:mtil-order1}, LoRA–Loop establishes a new state-of-the-art on order I, outperforming all baselines in transfer, average, and final accuracies. In particular, it preserves prior knowledge more effectively and gains an extra boost in the last task over GIFT, while also improving zero-shot transfer. On order II (Table~\ref{tab:mtil-order2}), our method also provides comprehensive improvements across every metric, confirming its robust balance of knowledge retention and new‐task adaptation.
\begin{table}[t]
  \centering
  \scriptsize
  \setlength{\tabcolsep}{3pt}
  \caption{Ablation study of different components. DST and AWC represent the distillation losses and AWC loss from GIFT. LFT and SF represent the LoRA-Finetuning and the Sample Filtering method in our framework.}
  \label{tab:ablation}
  \resizebox{\columnwidth}{!}{%
  \begin{tabular}{cc|cc|cc|cc|cc}
    \toprule
       +DST & +AWC & +LFT & +SF & Transfer & $\Delta$ & Avg. & $\Delta$ & Last & $\Delta$     \\
      \midrule
      \cmark &        &        &        & 68.9 & — & 76.6 & — & 85.0 & —  \\
      \cmark &        & \cmark &        & 68.7 & \textcolor{blue}{-0.2} & 76.8 & \textcolor{red}{+0.2} & 85.3 & \textcolor{red}{+0.3} \\
      \cmark &        &        & \cmark & 69.3 & \textcolor{red}{+0.4} & 76.8 & \textcolor{red}{+0.2} & 85.1 & \textcolor{red}{+0.1} \\ 
      \cmark &        & \cmark & \cmark & 69.0 & \textcolor{red}{+0.1} & 77.0 & \textcolor{red}{+0.4} & 85.9 & \textcolor{red}{+0.9} \\ 
      \midrule
      \midrule
      \cmark & \cmark &       &        & 69.7 & — &  77.3 & — & 85.4 & — \\
      \cmark & \cmark & \cmark&        & 69.8 & \textcolor{red}{+0.1} &  77.3 & 0.0 & 85.3 & \textcolor{blue}{-0.1} \\
      \cmark & \cmark &       & \cmark & 69.9 & \textcolor{red}{+0.2} & 77.4 & \textcolor{red}{+0.1} & 85.2 & \textcolor{blue}{-0.2} \\ 
      \cmark & \cmark & \cmark& \cmark & 69.8 & \textcolor{red}{+0.1} &  77.6 & \textcolor{red}{+0.3} & 86.0 & \textcolor{red}{+0.6} \\
  \bottomrule
  \end{tabular}
  }
\end{table}

\begin{table*}[t]
  \centering
  \setlength{\tabcolsep}{4pt}
  \caption{Analysis of important hyperparameters in our LoRA finetuning and sample filtering pipeline. Our default settings are marked in \colorbox{lightgray}{\strut gray}, while the best scores are marked in \textbf{bold}.}
  \label{tab:analysis}
  \begin{tabular}{@{}ccc@{}}
    \begin{subtable}[b]{0.33\textwidth}
      \centering
      \caption{LoRA rank $r$}
      \label{tab:analysis_a}
      \begin{tabular}{lccc}
        \toprule
        $r$    & Transfer & Avg.\ & Last \\
        \midrule
        2      & \textbf{69.8}     & 77.4  & 85.8 \\
        \rowcolor{lightgray}
        4      & \textbf{69.8}     & \textbf{77.6}  & \textbf{86.0} \\
        \rowcolor{white}
        8      & 69.7     & 77.4  & 85.9 \\
        16     & 69.6     & 77.3  & 85.6 \\
        \bottomrule
      \end{tabular}
    \end{subtable} &
    \begin{subtable}[b]{0.33\textwidth}
      \centering
      \caption{LoRA FT select num. $l$ (at $r=4$)}
      \label{tab:analysis_b}
      \begin{tabular}{lccc}
        \toprule
        $l$    & Transfer & Avg.\ & Last \\
        \midrule
        \rowcolor{lightgray}
        2      & \textbf{69.8}     & \textbf{77.6}  & \textbf{86.0} \\
        \rowcolor{white}
        4      & 69.7     & 77.5  & 85.9 \\
        8      & 69.7     & \textbf{77.6}  & 85.7 \\
        \bottomrule
      \end{tabular}
    \end{subtable} &
    \begin{subtable}[b]{0.33\textwidth}
      \centering
      \caption{LoRA FT select num. $l$ (at $r=16$)}
      \label{tab:analysis_c}
      \begin{tabular}{lccc}
        \toprule
        $l$    & Transfer & Avg.\ & Last \\
        \midrule
        2      & \textbf{69.6}     & \textbf{77.3}  & \textbf{85.6} \\
        4      & 69.3     & 77.1  & 85.4 \\
        8      & 69.5     & \textbf{77.3}  & \textbf{85.6} \\
        16     & 69.4     & 77.2  & \textbf{85.6} \\
        \bottomrule
      \end{tabular}
    \end{subtable} \\[4em]
    \begin{subtable}[b]{0.33\textwidth}
      \centering
      \caption{LoRA selection policy}
      \label{tab:analysis_d}
      \begin{tabular}{lccc}
        \toprule
        Policy   & Transfer & Avg.\ & Last \\
        \midrule
        \rowcolor{lightgray}
        Top \& Bottom & \textbf{69.8}     & \textbf{77.6}  & \textbf{86.0} \\
        \rowcolor{white}
        Random   & 69.4     & 77.2  & 85.6 \\
        Top      & 69.6     & 77.3  & 85.5 \\
        \bottomrule
      \end{tabular}
    \end{subtable} &
    \begin{subtable}[b]{0.33\textwidth}
      \centering
      \caption{Sampling selection policy}
      \label{tab:analysis_e}
      \begin{tabular}{lccc}
        \toprule
        Policy   & Transfer & Avg.\ & Last \\
        \midrule
        \rowcolor{lightgray}
        Top      & \textbf{69.8}     & \textbf{77.6}  & \textbf{86.0} \\
        \rowcolor{white}
        Middle & 69.6     & 77.3  & 85.4 \\
        Random   & 69.7     & 77.3  & 85.3 \\
        Bottom & 69.5     & 77.1  & 85.1 \\
        \bottomrule
      \end{tabular}
    \end{subtable} &
    \begin{subtable}[b]{0.33\textwidth}
      \centering
      \caption{Sampling budget $M_{pre}$}
      \label{tab:analysis_f}
      \begin{tabular}{lccc}
        \toprule
        $M_{pre}$    & Transfer & Avg.\ & Last \\
        \midrule
        2      &   69.7   & 77.3  & 85.5 \\
        4      &   69.6   & 77.5  & 85.9 \\
        \rowcolor{lightgray}
        8      & \textbf{69.8}     & \textbf{77.6}  & \textbf{86.0} \\
        \rowcolor{white}
        16     & \textbf{69.8}     & 77.5  & 85.9 \\
        \bottomrule
      \end{tabular}
    \end{subtable}
  \end{tabular}
  \vspace{0.3cm}
\end{table*}

\begin{figure*}[h!]
  \centering
   \includegraphics[width=\linewidth]{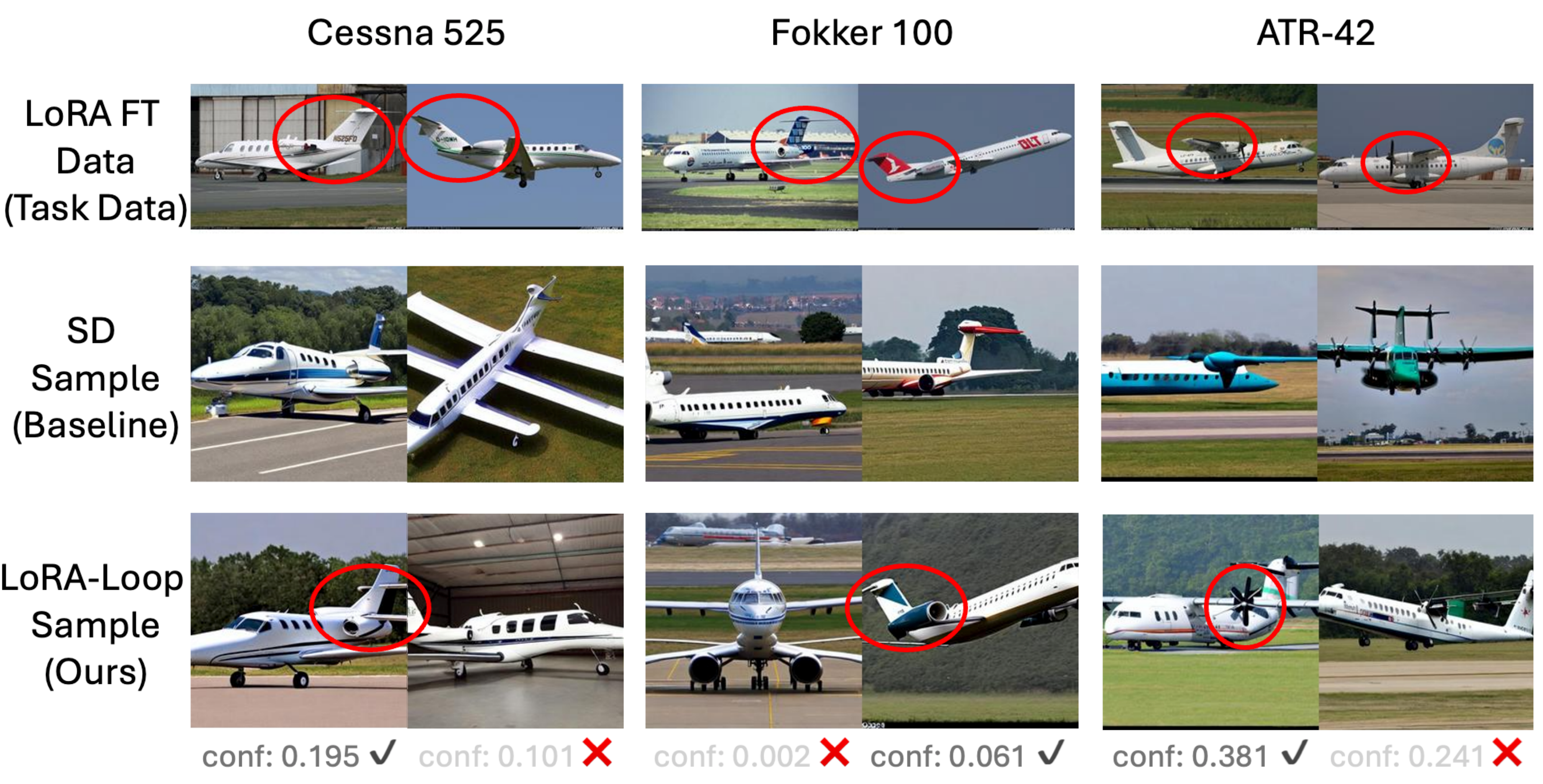}
   \caption{Visualization of the LoRA finetuning data and generation samples on Aircraft.}
   \label{fig:qual_aircraft}
\end{figure*}

\begin{figure*}[t]
  \centering
   \includegraphics[width=\linewidth]{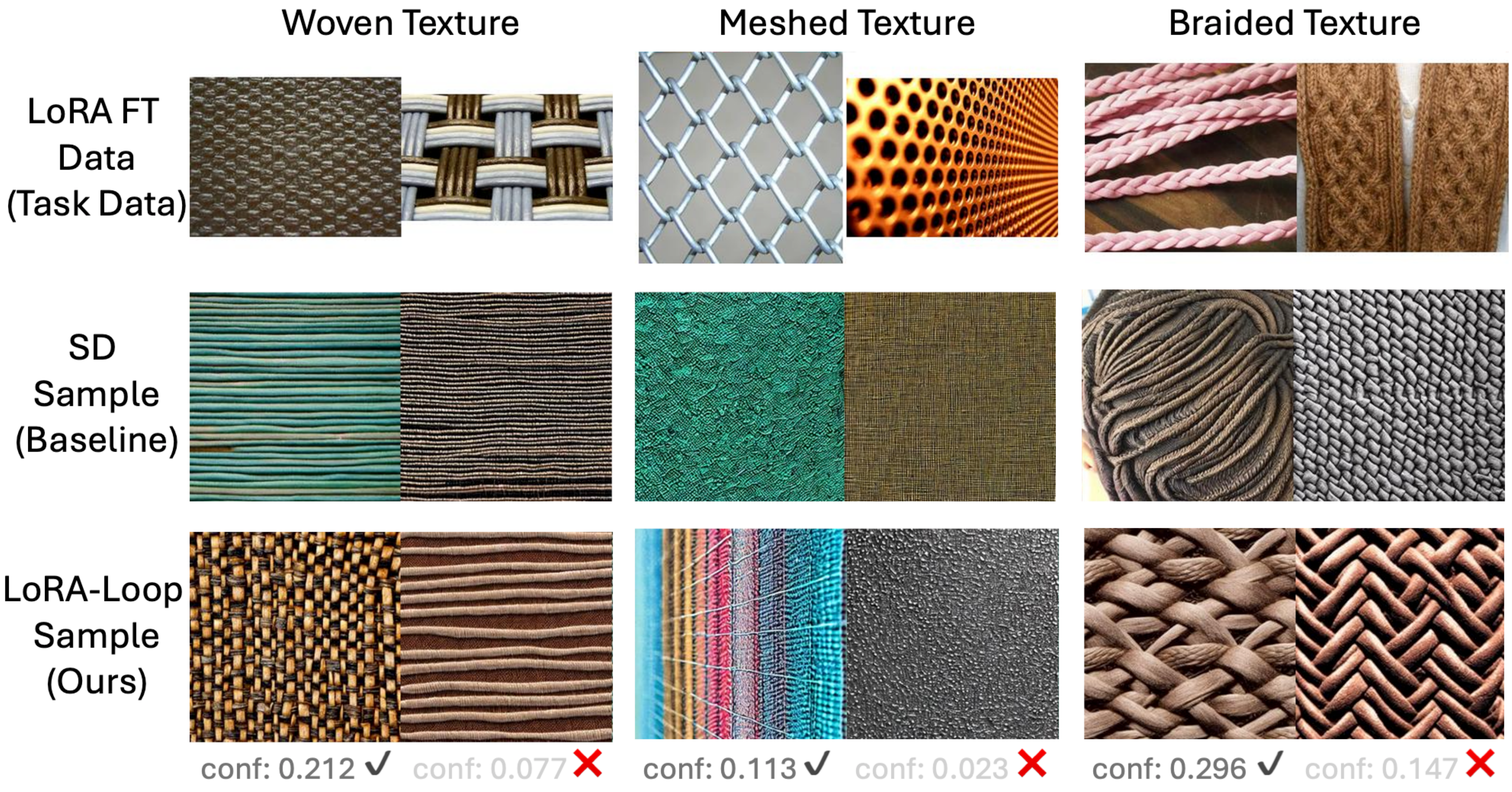}
   \vspace{-0.5cm}
   \caption{Visualization of the LoRA finetuning data and generation samples on DTD.}
   \label{fig:qual_texture}
\end{figure*}

\subsubsection{Ablation Studies}
\label{sec:ablation}
We evaluate each component of our method via an ablation study (Tab.~\ref{tab:ablation}), focusing on LoRA finetuning and our sample‐filtering step.  Since our framework builds on GIFT, which uses distillation losses ($\mathcal{L}_{CD}$ and $\mathcal{L}_{ITA}$), and a weight‐regularization loss ($\mathcal{L}_{AWC}$). To better demonstrate the improvement brought by our method to the distillation, we take “supervised + distillation” (i.e.\ without AWC) as our basic baseline, then report results both with and without the AWC loss.

The table shows that both LoRA finetuning and sample filtering yield consistent gains.  In particular, without AWC, adding LoRA + filtering boosts Avg. by 0.4 pp (vs.\ 0.3 pp with AWC) and Last by 0.9 pp (vs.\ 0.7 pp with AWC), which suggests our improvements concentrate on the distillation pipeline.
Further isolation of our two modules suggests that LoRA finetuning is particularly effective at preserving previously learned knowledge, evidenced by higher “Last” scores, whereas sample filtering more consistently maintains zero‐shot generalization, evidenced by higher “Transfer” scores. We believe this difference arises because filtering removes only the most obvious low‐quality outputs from Stable Diffusion but cannot rectify deeper domain or semantic misalignment, while LoRA Finetuning can close those subtler gaps at the risk of introducing occasional instability in generation quality during the finetuning. By combining both modules, our framework can simultaneously bridge domain and semantic gaps and stabilize post-finetuning outputs. Moreover, these components integrate seamlessly into the GIFT baseline and yield additional gains that push overall performance to a new state-of-the-art.

To assess hyperparameter sensitivity, we test five key settings and report results in Tab.~\ref{tab:analysis}: LoRA adapter rank $r$, number of per-class tuning examples $l$ at $r=4$ and $r=16$, LoRA training set selection policy, sample-filtering policy, and pre-filter sampling budget $M_{pre}$. As shown in Tab.~\ref{tab:analysis_a}, performance peaks at $r=4$, and increasing $r$ beyond causes a steady drop in all metrics, which we attribute to overly aggressive learning that degrades generation quality and thus weakens synthetic replay. With a moderate rank of $r=4$ (Tab.~\ref{tab:analysis_b}), just two examples per class suffice to reach peak alignment, whereas larger $l$ values yield marginal declines, likely because finite capacity cannot absorb too many samples. In the case of $r=16$ (Tab.~\ref{tab:analysis_c}), adding more tuning data neither improves stability nor boosts accuracy, suggesting that aggressive adaptation with limited prompt diversity can destabilize generation. 
Turning to selection policies, our Top \& Bottom training set selection scheme (Tab.~\ref{tab:analysis_d}) outperforms both random sampling and top-only confidence by inclusively covering both prototypical and edge cases during LoRA finetuning. In the sample filtering stage (Tab.~\ref{tab:analysis_e}), retaining only the highest-confidence generations delivers the best overall performance, while including mid- or low-confidence outputs noticeably degrades results. Finally, increasing the pre-filter budget (Tab.~\ref{tab:analysis_f}) steadily boosts performance up to $8$ samples per class, reflecting a more thorough search, but plateaus beyond that point. 
Overall, our experiments dissect each design choice, showing that LoRA finetuning and sample filtering jointly bridge domain and semantic gaps while stabilizing generation, achieving strong knowledge retention and zero-shot generalization across varied hyperparameters.

\subsection{Discussion}

\subsubsection{Qualitative Results}
\label{sec:qualitative}

To illustrate the efficacy of our domain/semantic alignment pipeline, we focus on two challenging MTIL datasets, Aircraft and DTD, where VLM's performance is relatively lower, as shown in our appendix. Figure~\ref{fig:qual_aircraft} presents three rows of images for three airplane types (Cessna 525, Fokker 100, ATR-42): the original task images used for LoRA finetuning, baseline samples from original Stable Diffusion, and outputs from our LoRA-finetuned model. Critical discriminative features, such as propeller design, tail shape, and engine placement, are often missed by the original Stable Diffusion model, resulting in visually plausible but incorrect generations, a clear manifestation of the \textit{semantic‐gap} issue. In contrast, the LoRA-finetuned model more faithfully reproduces these local attributes (e.g.\ the Cessna 525’s rear-mounted engine, the Fokker 100’s swept tail, and the ATR-42’s characteristic twin-prop assembly). Meanwhile, occasional low-quality outputs still occur, and our confidence-based filtering stage effectively suppresses these artifacts, ensuring that only high-quality samples proceed to the distillation step, ensuring the quality of synthetic replay.

Figure~\ref{fig:qual_texture} examines three texture classes, including woven, meshed, and braided, across diverse materials (fabric, cane, rope, and metal, etc.). Here, the main challenge is the wide variation in surface appearance, which is more related to a \textit{domain‐gap} issue. Original Stable Diffusion tends to collapse onto a narrow set of materials or patterns, whereas our LoRA-finetuned model generates textures that better match both the geometric pattern and the underlying substrate, as seen in the task data. As before, low-confidence or spurious generations are filtered out, yielding a synthetic replay set that preserves texture fidelity and material specificity.

\subsubsection{Comparison to Real Replay}
\label{sec:real_replay}
To validate our synthetic-replay approach and highlight its advantages, we compare against a VLM directly finetuned on the actual task images selected by LoRA-Loop for adapter training.  To isolate the effect of replay data on distillation, we disable the weight-regularization loss $\mathcal{L}_{AWC}$ and report the results in Tab.~\ref{tab:real_replay}.
Remarkably, using only two real examples per class, our synthetic-replay model nearly matches real-data replay while requiring only a fraction of the storage footprint.  As more real images are stored, storage costs grow linearly, and zero-shot transfer actually degrades slightly, indicating that direct image buffering can harm generalization in continual VLM learning.  By contrast, our approach preserves high accuracy and generalization, drastically reduces memory overhead, and avoids privacy risks associated with retaining real data

\begin{table}[t!]
  \centering
  \caption{Comparison to training with real replay data. Note that these results are obtained without AWC loss to study the effect of replay data on distillation.}
  \label{tab:real_replay}
  \resizebox{\columnwidth}{!}{%
  \begin{tabular}{l  c  c  c  c  c  c | c}
    \toprule
    Method & Transfer & $\Delta$ & Avg.\ & $\Delta$ & Last & $\Delta$ & Storage Cost \\
    \midrule
    GIFT~\cite{WuShWa25}& 68.9 & — & 76.6 & — & 85.0 & — & — \\
    Ours           & 69.0 & \textcolor{red}{+0.1} & 77.0 & \textcolor{red}{+0.4} & 85.9 & \textcolor{red}{+0.9} & 30.79 MB\\
    \midrule
    2 real replay/cls.            & 69.0 & \textcolor{red}{+0.1} & 77.6 & \textcolor{red}{+1.0} & 86.9 & \textcolor{red}{+1.9} & 118.95 MB\\
    4 real replay/cls.            & 68.9 & 0.0 & 78.0 & \textcolor{red}{+1.4} & 87.7 & \textcolor{red}{+2.7} & 189.34 MB\\
    8 real replay/cls.            & 68.8 & \textcolor{blue}{-0.1} & 78.2 & \textcolor{red}{+1.6} & 88.1 & \textcolor{red}{+3.1} & 327.94 MB\\
    \bottomrule
  \end{tabular}
  }
  \vspace{-0.2cm}
\end{table}


\section{Conclusion}
We introduce LoRA-Loop, a synthetic-replay framework for continual vision–language model learning that injects task-specific low-rank adapters into a frozen Stable Diffusion generator and employs a two-stage, confidence-based selection to align samples with the target task distribution for more effective replay distillation. Extensive experiments on the MTIL benchmark show that LoRA-Loop consistently outperforms prior synthetic-replay methods, achieving state-of-the-art transfer, average, and final accuracies, while preserving zero-shot generalization and reducing forgetting. Ablation studies and hyperparameter sweeps validate the robustness and impact of each design choice, and qualitative results highlight its ability to bridge both semantic and domain gaps. By seamlessly integrating generator adaptation into existing replay pipelines, LoRA-Loop offers a lightweight, privacy-preserving alternative to real-data buffering with minimal overhead.  
\clearpage

{
    \small
    \bibliographystyle{ieeenat_fullname}
    \bibliography{main}
}

\clearpage
\setcounter{page}{1}
\maketitlesupplementary

\section{Detailed Results}
Tab.~\ref{tab:transfer-all} and Tab.~\ref{tab:transfer-all-2} present the Detailed Transfer, Avg., and Last metrics for different continual‑training methods across the MTIL benchmark in Order I and Order II, respectively. These results highlight the ability of each method to adapt to new tasks while preserving knowledge learned from earlier ones.

In Order I (Tab.~\ref{tab:transfer-all}), our method achieves the best performance across most columns. Compared with other baselines, i.e., ZSCL, MoE‑Adapter, and GIFT, it delivers superior Transfer and Avg. metrics (69.8 vs. 69.7, 77.6 vs. 77.3), indicating its strong generalization across tasks. Its Last accuracy (86.0) also tops the chart, suggesting that it maintains the most robust performance after sequential training.
In Order II (Tab.~\ref{tab:transfer-all-2}), LoRA‑Loop similarly shows strong results across the Transfer, Avg., and Last metrics. Notably, it achieves the best Avg. (75.9) and Last (85.5) results, highlighting its ability to balance performance across both early and later tasks. Compared to other methods, LoRA‑Loop demonstrates better resistance to catastrophic forgetting and maintains higher overall performance across the varied domains and data shifts introduced by the different ordering of tasks.
These results collectively confirm that the proposed method maintains both strong plasticity for learning new tasks and high stability for preserving prior knowledge and zero-shot generalizability, making it highly effective across diverse and challenging continual VLM learning settings.

\newcommand{\rot}[1]{\rotatebox{30}{#1}}
\begin{table*}[h!]
  \centering
  \setlength{\tabcolsep}{3pt}
  \caption{Detailed Transfer, Avg., and Last accuracy (\%) of different continual‐training methods on the MTIL benchmark in Order I. $^*$ indicates reproduced results. The best score in each column is shown in \textbf{bold}.}
  \label{tab:transfer-all}

  \begin{subtable}[h!]{\textwidth}
  \centering
  \setlength{\tabcolsep}{3pt}
  \label{tab:transfer-detail}
  \begin{tabularx}{\textwidth}{l *{12}{C}}
    \toprule
    Method
      & \rot{Aircraft}
      & \rot{Caltech101}
      & \rot{CIFAR100}
      & \rot{DTD}
      & \rot{EuroSAT}
      & \rot{Flowers}
      & \rot{Food}
      & \rot{MNIST}
      & \rot{OxfordPet}
      & \rot{Cars}
      & \rot{SUN397}
      & \rot{Average} \\
    \midrule
    Zero-shot
      & 24.3 & 88.4 & 44.6 & 54.9 & 71.0 & 88.5 & 59.4 & 89.0 & 64.7 & 65.2 & 65.3 & 65.3 \\
    Fine-tune
      & 62.0 & 95.1 & 89.6 & 79.5 & 98.9 & 97.5 & 92.7 & 99.6 & 94.7 & 81.8 & 89.2 & 89.2 \\
    \midrule

    \multicolumn{13}{l}{\textbf{Transfer}} \\
    \midrule
    ZSCL~\cite{zheng2023preventing}
      &  & 86.0    & 67.4  & 45.4  & \textbf{50.4} & \textbf{71.0} & 87.6  & 61.8  & 86.8  & 60.1  & 66.8  & 68.1  \\
    MoE-Adapter~\cite{YuZ0H0LH24}
      &  & 87.9    & 68.2  & 44.4  & 49.9   & 70.7   & \textbf{88.7} & 59.7  & 89.1  & \textbf{64.5} & 65.5  & 68.9  \\
    GIFT$^*$~\cite{WuShWa25}
      &  & 88.2    & \textbf{69.9} & 46.3  & 48.8   & 69.8   & 87.3  & \textbf{69.2} & 89.0  & 59.9  & 68.1 & 69.7  \\
    LoRA-Loop (Ours)
      &  & \textbf{88.4} & 69.4  & \textbf{46.6} & 50.3   & 70.1   & 87.7  & 68.4  & \textbf{89.5} & 59.0  & \textbf{69.8} & \textbf{69.8} \\
    \midrule

    \multicolumn{13}{l}{\textbf{Avg.}} \\
    \midrule
    ZSCL~\cite{zheng2023preventing}
      & 45.1  & 92.0    & 80.1  & 64.3  & 79.5   & 81.6   & \textbf{89.6} & 75.2  & 88.4  & 64.7  & 68.0  & 75.4  \\
    MoE-Adapter~\cite{YuZ0H0LH24}
      & 50.2  & 91.9    & \textbf{83.1} & \textbf{69.4} & 78.9   & \textbf{84.0} & 89.1  & 73.7  & 89.3  & \textbf{67.7} & 66.9  & 76.7  \\
    GIFT$^*$~\cite{WuShWa25}
      & 50.9  & 93.7    & 80.9  & 67.3  & 79.8   & 83.6   & 89.3  & \textbf{80.1} & 90.5  & 64.7  & \textbf{69.3} & 77.3  \\
    LoRA-Loop
      & \textbf{52.2} & \textbf{95.0} & 81.2  & 67.5  & \textbf{80.5} & 83.7   & 89.5  & 79.6  & \textbf{90.8} & 64.0  & 69.2  & \textbf{77.6} \\
    \midrule

    \multicolumn{13}{l}{\textbf{Last}} \\
    \midrule
    ZSCL~\cite{zheng2023preventing}
      & 40.6  & 92.2    & 81.3  & 70.5  & 94.8   & 90.5   & \textbf{91.9} & 98.7  & 93.9  & 85.3  & 80.2  & 83.6  \\
    MoE-Adapter~\cite{YuZ0H0LH24}
      & 49.8  & 92.2    & \textbf{86.1} & \textbf{78.1} & 95.7   & \textbf{94.3} & 89.5  & 98.1  & 89.9  & 81.6  & 80.0  & 85.0  \\
    GIFT$^*$~\cite{WuShWa25}
      & 47.8  & 94.1    & 81.3  & 73.7  & 96.7 & \textbf{94.3} & 91.5  & \textbf{99.1} & \textbf{94.7} & 85.9  & 80.3  & 85.4  \\
    LoRA-Loop (Ours)
      & \textbf{50.7} & \textbf{96.5} & 81.8  & 74.4  & \textbf{96.9} & 94.1   & 91.5  & \textbf{99.1} & 94.4  & \textbf{86.2} & \textbf{80.4} & \textbf{86.0} \\
    \bottomrule
  \end{tabularx}
  \end{subtable}
\end{table*}

\begin{table*}[h!]
  \centering
  \setlength{\tabcolsep}{3pt}
  \caption{Detailed Transfer, Avg., and Last accuracy (\%) of different continual‐training methods on the MTIL benchmark in Order II. $^*$ indicates reproduced results. The best score in each column is shown in \textbf{bold}.}
  \label{tab:transfer-all-2}

  \begin{subtable}[h!]{\textwidth}
  \centering
  \setlength{\tabcolsep}{3pt}
  \label{tab:transfer-order2}
  \begin{tabularx}{\textwidth}{l *{12}{C}}
    \toprule
    Method
      & \rot{Cars}
      & \rot{Food}
      & \rot{MNIST}
      & \rot{OxfordPet}
      & \rot{Flowers}
      & \rot{SUN397}
      & \rot{Aircraft}
      & \rot{Caltech101}
      & \rot{DTD}
      & \rot{EuroSAT}
      & \rot{CIFAR100}
      & \rot{Average} \\
    \midrule
    Zero‐shot
      & 64.7 & 88.5 & 59.4 & 89.0 & 71.0 & 65.2 & 24.3 & 88.4 & 44.6 & 54.9 & 68.2 & 65.3 \\
    Fine‐tune
      & 89.6 & 92.7 & 94.7 & 97.5 & 97.5 & 81.8 & 62.0 & 95.1 & 79.5 & 98.9 & 89.6 & 89.2 \\
    \midrule

    \multicolumn{13}{l}{\textbf{Transfer}} \\
    \midrule
    ZSCL~\cite{zheng2023preventing}
      &   & 88.3   & 57.5   & 84.7   & 68.1   & 64.8   & 21.1   & 88.2   & 45.3   &55.2 & 68.2   & 64.2   \\
    MoE‐Adapter~\cite{YuZ0H0LH24}
      &   & \textbf{88.8} & 59.5   & 89.1   & 69.9   & 64.4   & 18.1   & 86.9   & 43.7   & 54.6   & 68.2   & 64.3   \\
    GIFT$^*$~\cite{WuShWa25}
      &   & 88.3   & 64.2   & 88.9   & \textbf{70.4} & 68.2 & 22.5   & 90.1   & 46.2   & 52.8   & 69.1   & 66.1   \\
    LoRA‐Loop (Ours)
      &   & 88.4   & \textbf{65.4} & \textbf{89.5} & 70.3   & \textbf{68.5}   & \textbf{23.3} & \textbf{90.4} & \textbf{47.1} & \textbf{69.4} & \textbf{69.4} & \textbf{66.3} \\
    \midrule
    \multicolumn{13}{l}{\textbf{Avg.}} \\
    \midrule
    ZSCL~\cite{zheng2023preventing}
      & 81.7   & \textbf{91.3} & 91.1   & 91.0   & 82.9   & 72.5   & 33.6   & 89.7   & 53.3   & \textbf{62.8} & 69.9   & 75.4   \\
    MoE‐Adapter~\cite{YuZ0H0LH24}
      & \textbf{84.9} & 89.9   & 89.3   & 91.4   & \textbf{86.2} & 72.2   & 33.4   & 89.4   & 53.3   & 61.4   & 69.9   & 74.7   \\
    GIFT$^*$~\cite{WuShWa25}
      & 83.5   & 91.0   & 92.7   & 93.1   & 85.9   & 74.4 & 35.7   & 92.0   & 54.4   & 60.8   & 70.7   & 75.8   \\
    LoRA‐Loop (Ours)
      & 83.3   & 91.1   & \textbf{92.9} & \textbf{93.3} & 86.1   & \textbf{74.6}   & \textbf{36.6} & \textbf{92.1} & \textbf{54.8} & 59.5   & \textbf{70.9} & \textbf{75.9} \\
    \midrule

    \multicolumn{13}{l}{\textbf{Last}} \\
    \midrule
    ZSCL~\cite{zheng2023preventing}
      & 78.2   & \textbf{91.1} & 97.6   & 92.5   & 87.4   & 78.2   & 25.0   & 92.3   & 72.7   & 96.2   & 86.3   & 83.4   \\
    MoE‐Adapter~\cite{YuZ0H0LH24}
      & \textbf{84.1} & 88.5   & 94.0   & 91.8   & \textbf{94.1} & 77.8 & 50.4   & 93.3   & \textbf{77.1}   & 87.7   & \textbf{86.6} & 84.1   \\
    GIFT$^*$~\cite{WuShWa25}
      & 81.1   & 90.3   & 98.6   & 94.2   & 91.7   & 78.8   & 50.8   & \textbf{94.4} & 75.5   & 95.3   & \textbf{86.6}   & 85.2   \\
    LoRA‐Loop (Ours)
      & 81.1   & 90.5   & \textbf{98.7} & \textbf{94.3} & 92.9 & \textbf{79.1} & \textbf{52.6} & 93.9   & 74.8 & \textbf{96.4} & 86.5   & \textbf{85.5} \\
      \bottomrule
  \end{tabularx}
  \end{subtable}
\end{table*}

\end{document}